\documentclass{article}

\usepackage{arxiv}

\usepackage[utf8]{inputenc} 
\usepackage[T1]{fontenc}    
\usepackage{hyperref}       
\usepackage{url}            
\usepackage{booktabs}       
\usepackage{amsfonts}       
\usepackage{nicefrac}       
\usepackage{microtype}      
\usepackage{lipsum}		
\usepackage{graphicx}
\usepackage{doi}

\usepackage{chemformula}

\usepackage{algorithm}
\usepackage{algpseudocode}
\usepackage{microtype}
\usepackage{graphicx}
\usepackage{multirow}
\usepackage{stfloats}
\usepackage[numbers]{natbib}

\usepackage{chngcntr}

\counterwithout{table}{section}

\title{LeMo-NADe: Multi-Parameter Neural Architecture Discovery with LLMs}

\author{Md Hafizur Rahman \\
	Department of Electrical and Computer Engineering\\
	University of Maine\\
	Orono, ME 04469 \\
	\texttt{md.hafizur.rahman@maine.edu} \\
	\And Prabuddha Chakraborty \\
	Department of Electrical and Computer Engineering\\
	University of Maine\\
	Orono, ME 04469 \\
	\texttt{prabuddha@maine.edu} \\
}

\renewcommand{\shorttitle}{}

\hypersetup{
pdftitle={LeMoNADe},
pdfsubject={q-bio.NC, q-bio.QM},
pdfauthor={Md Hafizur Rahman, Prabuddha Chakraborty},
pdfkeywords={Generative AI, Edge Systems, Large Language Model},
}

\begin{document}
\maketitle

\begin{abstract}
Building efficient neural network architectures can be a time-consuming task requiring extensive expert knowledge. This task becomes particularly challenging for edge devices because one has to consider parameters such as power consumption during inferencing, model size, inferencing speed, and \ch{CO2} emissions. In this article, we introduce a novel framework designed to automatically discover new neural network architectures based on user-defined parameters, an expert system, and an LLM trained on a large amount of open-domain knowledge. The introduced framework (LeMo-NADe) is tailored to be used by non-AI experts, does not require a predetermined neural architecture search space, and considers a large set of edge device-specific parameters. We implement and validate this proposed neural architecture discovery framework using CIFAR-10, CIFAR-100, and ImageNet16-120 datasets while using GPT-4 Turbo and Gemini as the LLM component. We observe that the proposed framework can rapidly (within hours) discover intricate neural network models that perform extremely well across a diverse set of application settings defined by the user.  
\end{abstract}

\section{Introduction}
Neural networks have found extensive application across various fields such as healthcare \cite{chua2023tackling,bhardwaj2023enhanced, nandy2023intelligent}, surveillance \cite{jaafar2023multimodal, mahum2023robust}, Industry 4.0 \cite{jan2023artificial, raja2023industry,shafiq2023continuous}, and Internet of Things (IoT) \cite{rajput2023fault,liyakat2023machine,thakkar2023attack}. A neural network can be composed of a large number of layers of different types while sporting diverse hyperparameters. Hence, for a given task: (1) finding the right set of neural layers; (2) connecting them in the right topology; and (3) selecting the most optimal hyperparameters for each layer can be a daunting task requiring a large amount of computation resources, human expert involvement, and time. Requiring a given neural network to perform under specific resource-constrained conditions (a case for many IoT/Edge devices) can add to the complexity of the neural architecture search process. For example designing a neural network to have more than $x\%$ accuracy for a given task is a hard problem to solve but it becomes harder if we further constrain the problem with additional parameters such as frames-per-second requirements during inferences and power consumption limits. 

Traditional neural architecture search (NAS) frameworks are typically designed to identify the optimal architecture within a specified search space. This approach, although powerful, is limited by the said pre-defined search space failing to innovate and create wildly new architectures. Traditional NAS techniques are designed to primarily focus on improving task accuracy with very little emphasis on additional parameters such as frames-per-second (FPS), power consumption, and green house gas emission that have become more relevant in recent years.

To mitigate these concerns we propose a \textbf{L}arge Languag\textbf{e} \textbf{Mo}del guided \textbf{N}eural \textbf{A}rchitecture \textbf{D}iscov\textbf{e}ry (LeMo-NADe) framework that can allow the discovery of novel neural network architecture without relying on a pre-defined search space. This will be achieved through an iterative approach utilizing a large language model (LLM) and an expert system for driving the LLM towards the target discovery. The expert system will use a set of configurable rules and several user-defined metrics to generate a set of instructions for the LLM leading to progressive refinement of the generated neural architecture.

To validate the LeMo-NADe framework, we perform extensive experimentation using the CIFAR-10, CIFAR-100, and ImageNet16-120 datasets. We use the framework to generate many different neural networks for diverse application requirements and priorities. Neural networks generated using LeMo-NADe for CIFAR-10 ($89.41\%$ test accuracy) and CIFAR-100 ($67.90\%$ test accuracy) showed near state-of-the-art level performance. For ImageNet16-120 LeMo-NADe was also able to generate fairly competitive architectures ($31.02\%$ test accuracy). LeMo-NADe is also very efficient (time, energy consumption, and \ch{CO2} Emissions) in terms of model generation/training. While using GPT-4 Turbo\footnote{\href{https://openai.com/pricing}{gpt-4-1106-preview}} \cite{openai2023gpt} (as the backend LLM), LeMo-NADe was able to generate and train CIFAR-100 models in about 4.81 hours consuming only about 0.50 kWh-PUE energy. LeMo-NADe is also capable of prioritizing other metrics (besides accuracy) towards creating neural architecture that are optimal for different IoT/Edge requirements (e.g. high speed low accuracy inferencing) and more importantly, has created novel neural architectures from scratch creating a new opportunity for search-space agnostic neural architecture search research.

To summarize, we offer the following contributions:
\begin{enumerate}
    \item Formalize and design a search-space agnostic neural architecture discovery framework (LeMo-NADe) leveraging Large Language Models (LLMs).
    \item Propose an expert system with associated rules and relevant metrics that is capable of driving a given LLM towards discovering different neural architectures.  
    \item Implement LeMo-NADe as a highly configurable/efficient tool for immediate application and easy future extensions.
    \item Qualitatively and quantitatively evaluate LeMo-NADe using CIFAR-10, CIFAR-100, and ImageNet16-120 datasets for diverse settings and application requirements. 
    
\end{enumerate}

\section{Background and Motivation}
In this section, we briefly describe the recent advances in the domain of neural architecture search. We also highlight the motivation behind our work stemming from different shortcomings of traditional NAS frameworks.  

\subsection{Neural Architecture Search}
Methods of Neural Architecture Search (NAS)  extensively applied across various research areas, such as image processing \cite{wang2023efficient,yang2023om,yang2023trustworthy}, signal processing \cite{dong2023rd,wang2023dymc,li2023graph}, object detection \cite{yuan2023ssob,jia2023fast}, and natural language processing\cite{mehta2023natural,girdhar2023benchmarking}. It involves identifying the best neural network (judged traditionally based on only accuracy) for a given task through repeated trials.
The early NAS techniques worked mainly based on the evolutionary algorithms (EA) \cite{real2019regularized} and reinforcement learning (RL) \cite{liu2018progressive}. Although these methods showed promising result by building quality networks they require high computing power and time. To solve this issue, weight-reusing \cite{cai2018efficient} approaches were proposed that avoids the necessity of training each design from the beginning resulting in low computation cost. One-shot approaches for NAS \cite{pham2018efficient} were also proposed which involves training a large network called SuperNet that incorporates every conceivable architecture within the search domain.
DNAS \cite{liu2018darts} is another weight re-using approach where all the SubNet parameters are optimized by gradient decent.

NAS is hard to reproduce due to its high computational power and time. To limit this issue researchers proposed NAS benchmark dataset that contains all the possible architecture with corresponding evaluation results. One NAS dataset is NAS-Bench-101 \cite{ying2019bench} that contains 5 million distinct neural architectures and is trained on CIFAR-10 dataset. The NAS-Bench-201 \cite{dong2020bench} dataset has 15625 cell layouts and is derived from a cell-based search technique and is trained on CIFAR-10 \cite{krizhevsky2009learning}, CIFAR-100 \cite{krizhevsky2009learning} and ImageNet16-120 \cite{chrabaszcz2017downsampled} datasets. In \cite{ye2022beta}, the authors proposed a NAS method named as $\beta$-DARTS to solve weak generalization ability found in DARTS method. They used the NAS-Bench-201 to evaluate their framework. In another research work \cite{movahedi2022lambda}, authors suggested $\Lambda$-DARTS as a solution for the structural flaws caused by the weight sharing approach in DARTS. In a recent work \cite{zheng2023can}, authors proposed a framework named GENIUS where they used LLM to solve the NAS problem while utilizing pre-defined search spaces.

\subsection{Shortcomings of NAS}

Traditional NAS algorithms typically focuses on improving accuracy and give little priority to parameters such as inferencing speed, training power consumption, and \ch{CO2} emissions. However, in recent years, it has become important to consider these parameters during the neural network design process due to: (1) an increase in the number of resource-constrained edge devices; and (2) environmental concerns arising from excessive AI-related power utilization and \ch{CO2} emissions. Most traditional NAS techniques also rely on having access to a pre-defined search space (of potential neural architectures) making it difficult to scale across different applications and use cases. 


\subsection{Why LLM for Neural Discovery?}
We hypothesize that a Large Language Model (LLM) that is trained on a large volume of open-domain data will inevitably also have the knowledge about different neural architectures. LLMs have demonstrated success in terms of searching for neural network architectures given a search space \cite{achiam2023gpt, zheng2023can, wang2023graph}. However, we wanted to go one step further and find out if LLMs can generate novel neural network architecture (discovery) without being pointed to a pre-defined search space. We also wanted to analyze if: (1) the open-domain knowledge has provided these LLMs with insights into different metrics associated with a neural architecture such as estimated training power consumption, inferencing speed, and \ch{CO2} emissions during inference; (2) these LLMs can follow automated instructions generated from an expert system for refining a neural network. 
\begin{figure}[h]
\begin{center}
\centerline{\includegraphics[width=0.7\columnwidth]{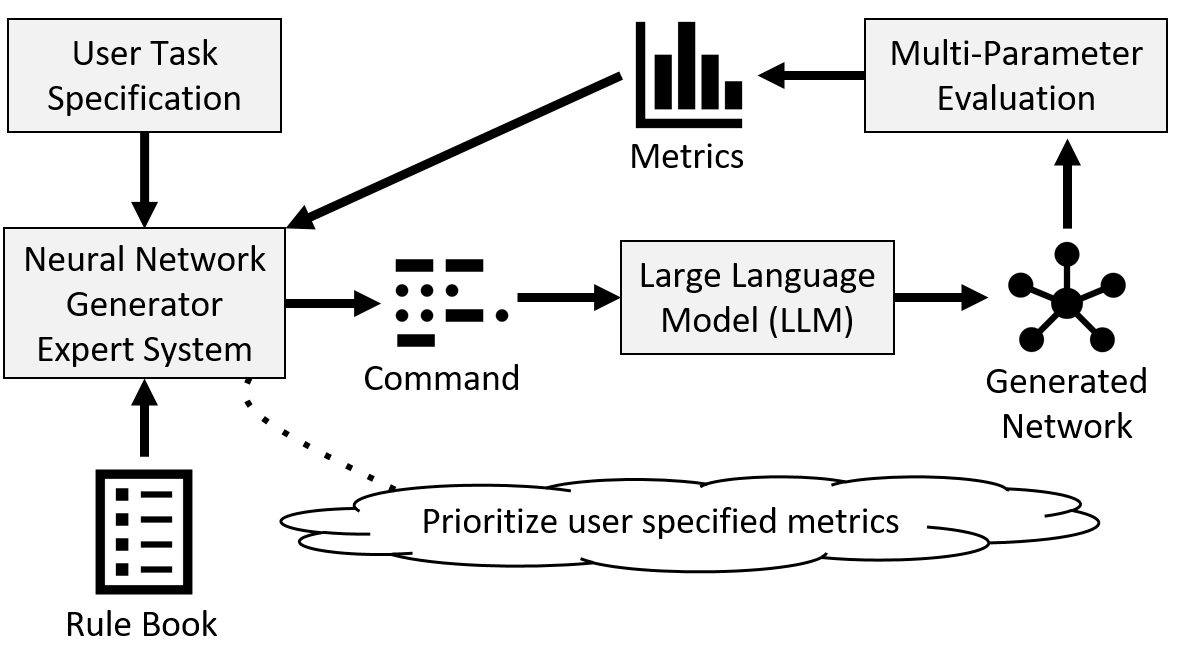}}
\caption{Overview of the LeMo-NADe framework.}
\label{overview}
\end{center}
\end{figure}

\section{Methodology}

In this section, we will describe and discuss the LeMo-NADe framework.

\subsection{Neural Discovery Process}

In Figure \ref{overview} we show an overview of the LeMo-NADe framework where an Expert System (ES) takes the task specification from the user (metrics) and generates commands for the LLM using a set of rules for creating a neural architecture. The generated neural network from the LLM is then evaluated and the associated evaluation-based metrics are used by ES to generate the next LLM command.

\begin{algorithm}[ht]

\caption{LeMo-NADe Neural Network Discovery}
\begin{algorithmic}[1]

\Function{AI\_AG}{$UDM$, $TS$, $TC, Initial\_Cmd$}
    \State $BCM \gets \emptyset$
    \State $Metrics \gets UDM$
    \State $BestModel \gets \emptyset$
    \State $cmd \gets Initial\_Cmd$ 
    \State $TC=False$
    \While{$TC=False$}
        \State Response $\gets$ LLM\_AGI(cmd)
        \State Model $\gets$ Create\_Model(Response)
        \State Metrics.P $\gets$ Model.params
        \State [Model, $Metrics.E_1$] $\gets$ Model.train(TS.train\_set)
        \State $Metrics.A_1$ $\gets$ Model.eval(train\_set)
        \State [$Metrics.E_2,Metrics.A_2$] $\gets$ Model.eval(val)
        \State cmd $\gets$ NNGES($Metrics$)
        \State cmd $\gets$ RESOLVE\_CONFLICT(cmd)

        \State $CM \gets Model\_Effectiveness(Metrics)$
        \If{$CM>BCM$}
        \State BCM $\gets$ CM
        \State BestModel $\gets$ Model
        
        \EndIf
        
        \If{Termination Condition is Met}
            \State $TC=True$
        \EndIf
        
    \EndWhile
\State \Return BestModel
\EndFunction
\end{algorithmic}
\label{algo:AI_NAS}
\end{algorithm}

\begin{table*}[ht]
\centering
\begin{center}
\begin{small}
\begin{sc}
\caption{Proposed user-defined and evaluation based metrics.}
\label{tab:Metrics}
\begin{tabular}{lcccr}
\toprule
Legends & Description & Type              \\ 
\midrule
$A_1$    & Training accuracy of the current model (CM) predicted by LLM/AGI     &   Evaluation based    \\ 
$PA_1$ & Priority of the Training Accuracy & User defined \\
$TA_1$ & Threshold of the Training accuracy & User defined \\ 

$A_2$ & Validation accuracy of the current model (CM) predicted by LLM/AGI    &      Evaluation based        \\ 
$PA_2$ & Priority of the Validation Accuracy &  User defined \\
$TA_2$ & Threshold of the Validation accuracy &  User defined \\ 

$E_1$ & Energy required for evaluating the training set    &        Evaluation based      \\
$PE_1$ & Priority of the Energy required for evaluating the training set & User defined  \\ 
$TE_1$ & Threshold of Energy required for evaluating the training set &  User defined \\ 

$E_2$ & Energy required for evaluating the validation set    &      Evaluation based        \\ 
$PE_2$ & Priority of the Energy required for evaluating the validation set & User defined  \\ 
$TE_2$ & Threshold of Energy required for evaluating the validation set & User defined  \\ 
$F$ & FPS of the current model (CM) predicted by LLM/AGI  &     Evaluation based         \\ 
$PF$ & Priority of the FPS for the model &  User defined \\ 
$TF$ & Threshold ofthe FPS for the model & User defined  \\ 

$P$ & Parameters of the current model (CM) predicted by LLM/AGI    &      Evaluation based        \\ 

$OT$ & Threshold value to check the Overfitting &  User defined \\ 
$UT$ & Threshold value to check the Underfitting  &  User defined \\ 
\bottomrule
\end{tabular}
\end{sc}
\end{small}
\end{center}
\end{table*}

Algorithm \ref{algo:AI_NAS} shows the overall procedure of LeMo-NADe. It takes user-defined metrics (UDM), task specification (TS), termination condition (TC), and the initial command ($Initial\_Cmd$) as input from the user. From lines 2-6, all variables are initialized and the initial command (designed based on the task specification) is stored in the cmd variable. In line 8, LLM\_AGI (the GPT-4 Turbo \cite{openai2023gpt} and Gemini \cite{team2023gemini} as LLM component) returns a response based on the command and subsequently, a model is created from the response. From lines 11-13, the generated model is trained and evaluated with the training and validation datasets respectively and a set of metrics such as training energy ($E_1$), training accuracy ($A_1$), validation energy ($E_2$) and validation accuracy ($A_2$) are calculated and stored in the Metrics dictionary. The Metrics dictionary is then passed to the NNGES (Algorithm \ref{algo:NNGES}) to generate a set of instructions for the next round of LLM-based neural network generation. Any conflicts between the generated instructions are removed using Algorithm \ref{algo:RC}. To store the best network architecture, we calculate a metric called the combined model effectiveness ($CM$) which we define as: 

\begin{equation}
CM = AW \cdot (T_a + V_a) + (FW \cdot NF) - EW \cdot (T_{NE} + V_{NE}) 
\end{equation}

Where, $AW$ is weight for accuracy; $T_a$ is training accuracy; $V_a$ is validation accuracy; $FW$ is weight for FPS; $NF$ is normalized FPS; $EW$ is weight for energy; $T_{NE}$ is normalized training energy; $V_{NE}$ is normalized validation energy. Table \ref{tab:Metrics} describes all metrics.

\begin{algorithm}[ht]
\caption{LeMo-NADe Expert System}
\scriptsize\addtolength{\tabcolsep}{1.5pt}
\renewcommand{\arraystretch}{1}
\begin{algorithmic}[1]
\Function{NNGES}{$Metrics$}

\State $cmd \gets \{ACL: 0, ASC: 0, ADL: 0, RCL: 0, RSC: 0, RDL: 0, AD: 0, AMK: 0, AWI: 0, AR: 0, RK: 0, RD: 0, AMN: 0, RN: 0, RR: 0\}$

\If {$Metrics.A_1 < Metrics.TA_1$}  
        \State $cmd[ACL] \gets cmd[ACL] + Metrics.PA_1$ 
        \State $cmd[ADL] \gets cmd[ADL] + Metrics.PA_1$ 
        \State $cmd[AMK] \gets cmd[AMK] + Metrics.PA_1$
        \State $cmd[ASC] \gets cmd[ASC] + Metrics.PA_1$
\EndIf
    
\If {$Metrics.A_2<Metrics.TA_2$} 
        \State $cmd[ACL]  \gets cmd[ACL] + Metrics.PA_2$ 
        \State $cmd[ADL]  \gets cmd[ADL] + Metrics.PA_2$ 
        \State $cmd[AMK]  \gets cmd[AMK] + Metrics.PA_2$ 
        \State $cmd[ASC]  \gets cmd[ASC] + Metrics.PA_2$
\EndIf

\If {$Metrics.E_1>Metrics.TE_1$} 
        \State $cmd[AD]  \gets cmd[AD] + Metrics.PE_1$
        \State $cmd[AWI]  \gets cmd[AWI] + Metrics.PE_1$
        \State $cmd[RCL]  \gets cmd[RCL] + Metrics.PE_1$
        \State $cmd[RK]  \gets cmd[RK] + Metrics.PE_1$
        \State $cmd[RSC]  \gets cmd[RSC] + Metrics.PE_1$
\EndIf

\If {$Metrics.E_2>Metrics.TE_2$} 
        \State $cmd[AD]  \gets cmd[AD] + Metrics.PE_2$
        \State $cmd[RCL]  \gets cmd[RCL] + Metrics.PE_2$
        \State $cmd[RK]  \gets cmd[RK] + Metrics.PE_2$
        \State $cmd[RDL]  \gets cmd[RDL]+ Metrics.PE_2$
\EndIf

\If {$Metrics.F<Metrics.TF$} 
        \State $cmd[AD]  \gets cmd[AD] + Metrics.PF$
        \State $cmd[RK]  \gets cmd[RK] + Metrics.PF$
        \State $cmd[RSC]  \gets cmd[RSC] + Metrics. PF$
\EndIf

\If {$(Metrics.A_1-Metrics.A_2)>Metrics.OT$} 
        \State $cmd[ADL]  \gets cmd[ADL] + Metrics.PA_1$
        \State $cmd[RCL]  \gets cmd[RCL] + Metrics.PA_1$
        \State $cmd[RK]  \gets cmd[RK] + Metrics.PA_1$
        \State $cmd[RN]  \gets cmd[RN] + Metrics.PA_1$
        \State $cmd[AR]  \gets cmd[AR] + Metrics.PA_1$
\EndIf

\If {$(Metrics.A_2-Metrics.A_1)>Metrics.UT$} 
        \State $cmd[RD]  \gets cmd[RD] + Metrics.PA_2$
        \State $cmd[ACL]  \gets cmd[ACL] + Metrics.PA_2$
        \State $cmd[AMK]  \gets cmd[AMK] + Metrics.PA_2$
        \State $cmd[ADL]  \gets cmd[ADL] + Metrics.PA_2$
        \State $cmd[AMN]  \gets cmd[AMN] + Metrics.PA_2$
        \State $cmd[ASC]  \gets cmd[ASC] + Metrics.PA_2$
        \State $cmd[RR]  \gets cmd[RR] + Metrics.PA_2$
\EndIf
\State \Return{$cmd$}
\EndFunction 
\end{algorithmic}
\label{algo:NNGES}
\end{algorithm}

\begin{table}[ht]
\centering
\begin{center}
\begin{small}
\begin{sc}
\caption{Rule Book for the LeMo-NADe Expert System.}
\label{tab:instSet}
\renewcommand{\arraystretch}{1}
\small\addtolength{\tabcolsep}{-5pt}
\begin{tabular}{lcccr}
\toprule
\textbf{Legend} & \textbf{Description}  &\textbf{Conflicting Instructions} \\
\midrule
acl    & Add Convolutional Layer    & rcl         \\ 
asc    & Add Skip Connection        & rsc         \\
adl    & Add Dense Layer            & rdl         \\ 
rcl    & Reduce Convolutional Layer & acl         \\ 
rsc    & Reduce Skip Connection     & asc         \\ 
rdl    & Reduce Dense Layer         & adl         \\ 
ad     & Add Dropout Layer          & rd           \\ 
amk    & Add More Kernel            & rk          \\ 
awi    & Add Weight Initializer     & -           \\ 
ar     & Add Regularization         & rr           \\ 
rk     & Reduce Number of Kernel    & amk         \\ 
rd     & Reduce Dropout Layer       & ad         \\ 
amn    & Add More Neurons           & rn          \\
rn     & Reduce Neurons             & amn         \\
rr     & Reduce Regularization      & ar          \\ 
\bottomrule
\end{tabular}
\end{sc}
\end{small}
\end{center}
\end{table}

\subsection{Expert System for Instruction set Generation}
\label{sub:ES}

The expert system drives the LLM towards constructing an optimal neural network for a given set of user-defined parameters by generating a set of instructions based on the full set of Metrics. Algorithm \ref{algo:NNGES} shows the overall process for instruction generation. 
Table \ref{tab:instSet} describes all instructions used by ES. In line 3, if the $A_1$ (training accuracy) is below the threshold value as defined by the user then we add the associated priority value (for this metric) to certain instructions inside $cmd$. This process leads to the selection and prioritization of instructions that are used by Algorithm \ref{algo:AI_NAS} for generating the LLM command for the next iteration.  


\subsection{Conflict Resolution}
\label{sub:RC}
Instructions generation might have some conflicts (see Table \ref{tab:instSet}) due to the nature of the instructions themselves. Algorithm \ref{algo:RC} shows the overall procedure of eliminating conflicts. In line 3, we organize the cmd in descending order according to their values. Through lines 7-9, it identifies and stores the commands that have values larger than 0 in $Refined\_Cmd$. In lines 10-15, the algorithm examines the current command and the remaining commands to identify any conflicts. If a conflict is detected, the corresponding value is set to zero. 

\begin{algorithm}[H]
\caption{LeMo-NADe Conflict Resolution}
\small\addtolength{\tabcolsep}{1pt}
\begin{algorithmic}[1]
\Function{Resolve\_ Conflict}{cmd}
    \State $Refined\_Cmd \gets \emptyset$
    \State $\text{sort\_descending}(cmd)$
    \State $m \gets 0$
    \State $key \gets cmd.keys()$
    \While{$m < \text{Length}(cmd)$}
            \If{$cmd[key[m]] > 0$}
                \State $Refined\_Cmd[key[m]] \gets cmd[key[m]]$
                \State $n \gets m+1$
                \While{$n < \text{Length}(cmd)$}
                    \If{$cmd[key[m]].\text{conflict}=cmd[key[n]]$}
                        \State $cmd[key[n]] \gets 0$
                    \EndIf
                    \State $n \gets n + 1$
                \EndWhile
            \EndIf
            \State $m \gets m + 1$
    \EndWhile
    \State \Return{$Refined\_Cmd$}
\EndFunction
\end{algorithmic}
\label{algo:RC}
\end{algorithm}

\section{Experimental Analysis}
In this study, we have considered three publicly available datasets CIFAR-10 \cite{krizhevsky2009learning}, CIFAR-100 \cite{krizhevsky2009learning}, and ImageNet16-120 \cite{chrabaszcz2017downsampled} for validating LeMo-NADe. All experiments are run on a single NVIDIA A100 GPU (to make a fair determination of metrics such as power consumption and runtime).

\begin{table*}[ht]
\centering
\begin{center}
\begin{small}
\begin{sc}
\caption{Effectiveness of LeMo-NADe for generating optimal neural network models for CIFAR-10 Dataset.}
\label{tab:cifar10}
\renewcommand{\arraystretch}{1}
\small\addtolength{\tabcolsep}{-7pt}
\begin{tabular}{lccccccccr}
\toprule
\textbf{LLM} & \textbf{Setting} & \textbf{\begin{tabular}[c]{@{}c@{}}Test \\ Accuracy\end{tabular}} & \textbf{\begin{tabular}[c]{@{}c@{}}Number\\ of \\ Params\end{tabular}} & \textbf{\begin{tabular}[c]{@{}c@{}}Generating \\ + \\ Training\\ Runtime \\ (Hours)\end{tabular}} & \textbf{\begin{tabular}[c]{@{}c@{}}Generating \\ + \\ Training\\ Energy \\ (kWh-PUE)\end{tabular}} & \textbf{\begin{tabular}[c]{@{}c@{}}Generating \\ + \\ Training\\ \ch{CO2} \\ Emission \\ (lbs)\end{tabular}} & \textbf{\begin{tabular}[c]{@{}c@{}}Inferencing \\ Frame \\ Per \\ Second\end{tabular}} & \textbf{\begin{tabular}[c]{@{}c@{}}Inferencing\\ Energy \\ Per Image\\ (kWh-PUE)\end{tabular}} & \textbf{\begin{tabular}[c]{@{}c@{}}Inferencing\\ \ch{CO2} \\ Emission \\ (lbs)\end{tabular}} \\ 

\midrule
 & 1                & 82.55 \%                                                                & 2442698                                                                & 3.93                                                                                          & 0.5157                                                                                       & 0.4920                                                                                           & 6897.22                                                                               & 1.11E-05                                                                                      & 1.06E-05                                                                                \\ 
 & 2                & 84.27\%                                                                & 357706                                                                 & 4.60                                                                                          & 0.5682                                                                                       & 0.5421                                                                                          & 115220.68                                                                              & 5.34E-06                                                                                      & 5.09E-06                                                                                \\ 
 & 3                & 89.41 \%                                                                & 699466                                                                 & 4.25                                                                                          & 0.7041                                                                                       & 0.6717                                                                                          & 10755.78                                                                               & 7.06E-06                                                                                      & 6.74E-06                                                                                \\ 
 & 4                & 69.26 \%                                                                & 136874                                                                 & 4.84                                                                                          & 0.4751                                                                                       & 0.4533                                                                                         & 14636.72                                                                               & 5.39E-06                                                                                      & 5.14E-06                                                                                \\ 
\multirow{-5}{*}{\textbf{\begin{tabular}[c]{@{}c@{}}GPT-4 \\ Turbo\end{tabular}}} & 5                & 85.44 \%                                                                & 2400330                                                                & 4.27                                                                                          & 0.5743                                                                                       & 0.5480                                                                                          & 13051.73                                                                               & 5.89E-06                                                                                      & 5.62E-06                                                                                \\ 
\midrule

& 1                & 81.76 \%                                                                & 440778                                                                & 2.14                                                                                         & 0.1605                                                                                     & 0.1531                                                                                          & 13435.93                                                                              & 3.18e-06                                                                                     & 3.04e-06                                                                               \\ 
& 2                & 81.20 \%                                                                & 128842                                                                 & 2.79                                                                                        & 0.2333                                                                                      & 0.2226                                                                                           & 21553.23                                                                    & 2.44e-06                                                                                      & 2.33e-06                                                                                \\ 
& 3                & 80.79 \%                                                                & 3016522                                                                 & 2.403                                                                                         & 0.2064                                                                                  & 0.1969                                                                                         & 20460.87                                                                            & 1.08e-06                                                                                      &1.03e-06                                                                                \\ 
& 4                & 71.73 \%                                                                & 79530                                                                 & 2.87                                                                                       & 0.1811                                                                                      & 0.1728                                                                                           & 23029.86                                                                               & 7.42e-07                                                                                      & 7.08e-07                                                                               \\ 
\multirow{-5}{*}{\textbf{Gemini}} & 5                & 78.59 \%                                                                & 1276234                                                                & 1.94                                                                                        & 0.1264                                                                                       & 0.1206                                                                                           & 19565.53                                                                             & 6.42E-06                                                                                      &6.12E-06                                                                                \\ 
\bottomrule

\end{tabular}
\end{sc}
\end{small}
\end{center}
\end{table*}

\subsection{Search and Final Training Strategy}
Both the CIFAR-10 and CIFAR-100 datasets comprises 60k images, each with dimensions of $32\times32$ pixels where 50k and 10k samples are designated for training and testing purposes respectively. The CIFAR-10 dataset have 10 output classes where CIFAR-100 datasets have 100 output classes. We split the training data in a 9:1 ratio for generating the new training and validation set respectively. ImageNet16-120 has 151k training and 6k testing samples with a resolution of $16\times16$ having 120 classes. We split the training dataset again in a 9:1 ratio towards creating a new training and validation set. During neural discovery, we run LeMo-NADe for 30 iterations and then train the generated network for 20 epochs using batch size of 512 (Adam optimizer and 0.001 learning rate). The learning rate reduces by 10\% of its initial value if validation loss doesn't reduce for successive 5 epochs. To reduce overfitting, we augmented the training dataset with random rotation of 20 degrees, random horizontal flip, 10\% random width shift, 10\% random height shift, 10\% random shear, and 10\% random zoom. 

\begin{table}[ht]
\centering
\begin{center}
\begin{small}
\begin{sc}
\caption{User-defined parameters for different experiment settings.}
\label{tab:settings}
\renewcommand{\arraystretch}{0.9}
\small\addtolength{\tabcolsep}{1pt}
\begin{tabular}{cccccc}
\toprule
Setting & $PA_1$ & $PA_2$ & $PE_1$ & $PE_2$ & $PF$  \\
\midrule
1       & 0.4 & 0.6 & 0   & 0   & 0  \\ 
2       & 0.3 & 0.3 & 0.2 & 0.2 & 0   \\ 
3       & 0   & 1   & 0   & 0   & 0    \\ 
4       & 0   & 0   & 0.5 & 0.5 & 0  \\ 
5       & 0.3 & 0.4 & 0   & 0   & 0.3   \\ 

\bottomrule
\end{tabular}
\end{sc}
\end{small}
\end{center}
\end{table}

After LeMo-NADe determines the most optimal neural network, we retrain it for 500 epochs (max) using a batch size of 1024. The Adam optimizer was utilized, starting with an initial learning rate of 0.01. This learning rate is decreased if there is no reduction in validation loss over 20 consecutive epochs. We configured the GPT-4 Turbo with temperature setting of 0.

\subsection{Results}
\subsubsection{CIFAR-10 Dataset}
Table \ref{tab:cifar10} shows the experimental results for CIFAR-10 dataset (settings described in Table \ref{tab:settings}). 
For GPT-4 Turbo (Table \ref{tab:cifar10}), we obtain better test accuracy of 89.41\% when the validation accuracy priority is set to 1. For this case, the total run time (searching time + model training) is 4.25 hours, the power utilization effectiveness (PUE) is 0.7041 kWh , and \ch{CO2} emission is 0.6717 lb. Since both training and inferencing energy are prioritized in setting 4, LeMo-NADe generated a much simpler neural network while sacrificing some validation accuracy (69.26\%). 
We calculated the PUE and \ch{CO2} emission using the following two equations as described in \cite{strubell2019energy}.

\begin{equation}
p_t=\frac{1.58 t\left(p_c+p_r+g p_g\right)}{1000}
\label{eqn1}
\end{equation}

\begin{equation}
\ch{CO2} \mathrm{e}=0.954 p_t
\label{eqn2}
\end{equation}

Here, $p_c$, $p_r$, and $p_g$ represent the power usages (in watt) of CPU, RAM and GPU respectively. Also, t is the total run time in hours and g is the number of GPUs.

For Gemini, the highest test accuracy of 81.76\% was obtained for setting 1 where LeMo-NADe prioritised training accuracy and validation accuracy. This generation (and training) took 2.14 hours and consumed 0.1605 kWh-PUE of energy emitting 0.1531 lb of \ch{CO2}. The efficient model for inferencing (in terms of energy consumed) was obtained for setting 4 where LeMo-NADe gave equal priority to training and inferencing time energy consumption. For this setting, LeMo-NADe took 2.87 hours (generating and training time), consumed 0.1811 kWh-PUE, and emitted 0.1728 lb \ch{CO2}. We can see that GPT-4 Turbo perform better compared to Gemini.

\begin{table*}[ht]
\centering

\begin{center}
\begin{small}
\begin{sc}
\caption{Effectiveness of LeMo-NADe for generating optimal neural network models for CIFAR-100 Dataset.}
\label{tab:cifar100}
\renewcommand{\arraystretch}{1}
\small\addtolength{\tabcolsep}{-7pt}
\begin{tabular}{lccccccccr}
\toprule
\textbf{LLM} &\textbf{Setting} & \textbf{\begin{tabular}[c]{@{}c@{}}Test \\ Accuracy\end{tabular}} & \textbf{\begin{tabular}[c]{@{}c@{}}Number\\ of \\ Params\end{tabular}} & \textbf{\begin{tabular}[c]{@{}c@{}}Generating \\ + \\ Training\\ Runtime \\ (Hours)\end{tabular}} & \textbf{\begin{tabular}[c]{@{}c@{}}Generating \\ + \\ Training\\ Energy \\ (kWh-PUE)\end{tabular}} & \textbf{\begin{tabular}[c]{@{}c@{}}Generating \\ + \\ Training\\ \ch{CO2}  \\ Emission \\ (lbs)\end{tabular}} & \textbf{\begin{tabular}[c]{@{}c@{}}Inferencing \\ Frame \\ Per \\ Second\end{tabular}} & \textbf{\begin{tabular}[c]{@{}c@{}}Inferencing\\ Energy \\ Per Image\\ (kWh-PUE)\end{tabular}} & \textbf{\begin{tabular}[c]{@{}c@{}}Inferencing\\ \ch{CO2}  \\ Emission \\ (lbs)\end{tabular}} \\ 

\midrule
& 1                & 56.94\%                                                                 & 948708                                                                 & 4.46                                                                                          & 0.6874                                                                                       & 0.6558                                                                                                                & 11753.98                                                                               & 7.95E-06                                                                                      & 7.59E-06                                                                                \\ 
& 2                & 63.43\%                                                                 & 1725348                                                                & 4.73                                                                                          & 0.7142                                                                                       & 0.6813                                                                                                               & 10278.36                                                                               & 6.08E-06                                                                                      & 5.80E-06                                                                                \\ 
& 3                & 58.49\%                                                                 & 948708                                                                 & 4.13                                                                                          & 0.6056                                                                                       & 0.5778                                                                                                                & 13294.64                                                                               & 5.96E-06                                                                                      & 5.68E-06                                                                                \\ 
& 4                & 67.90\%                                                                 & 79530                                                                  & 4.81                                                                                          & 0.5021                                                                                       & 0.4790                                                                                                                & 15079.86                                                                               & 5.63E-06                                                                                      & 5.37E-06                                                                                \\
\multirow{-5}{*}{\textbf{\begin{tabular}[c]{@{}c@{}}GPT-4 \\ Turbo\end{tabular}}} & 5                & 64.55\%                                                                 & 3489316                                                                & 4.68                                                                                          & 0.7351                                                                                       & 0.7013                                                                                                                & 10026.15                                                                               & 7.44E-06                                                                                      & 7.10E-06                                                                                \\ 
\midrule
& 1                & 47.78 \%                                                                & 721060                                                                & 2.21                                                                                        & 0.1609                                                                                      & 0.1535                                                                                           & 21455.82                                                                             & 2.91e-06                                                                                  & 2.77e-06                                                                                \\ 
& 2                & 52.96 \%                                                                & 801444                                                                 & 2.48                                                                                         & 0.0932                                                                                      & 0.0889                                                                                           & 20740.8                                                                              & 2.48E-06                                                                                      & 2.36E-06                                                                                \\ 
& 3                & 50.20 \%                                                                & 4385540                                                                 & 2.65                                                                                         & 0.2877                                                                                       & 0.2744                                                                                           & 18524.72                                                                              & 2.67E-06                                                                                      & 2.55E-06                                                                                \\ 
& 4                & 10.24 \%                                                                & 12836                                                                 & 2.83                                                                                         & 0.1015                                                                                       & 0.097                                                                                           & 20726.49                                                                              & 1.19E-06                                                                                      & 2.09E-06                                                                                \\ 
\multirow{-5}{*}{\textbf{Gemini}}& 5                & 51.33 \%                                                                & 2665124                                                                & 2.92                                                                                          & 0.2115                                                                                       & 0.2018                                                                                           & 18470.16                                                                               & 1.04E-06                                                                                      & 9.90E-07                                                                                \\ 
\bottomrule
\end{tabular}
\end{sc}
\end{small}
\end{center}
\end{table*}
\begin{table*}[ht]
\centering
\begin{center}
\begin{small}
\begin{sc}
\caption{Effectiveness of LeMo-NADe for generating optimal neural network models for ImageNet16-120 Dataset.}
\label{tab:ImageNet}
\renewcommand{\arraystretch}{1}
\small\addtolength{\tabcolsep}{-7pt}
\begin{tabular}{lccccccccr}
\toprule
\textbf{LLM} &\textbf{Setting} & \textbf{\begin{tabular}[c]{@{}c@{}}Test \\ Accuracy\end{tabular}} & \textbf{\begin{tabular}[c]{@{}c@{}}Number\\ of \\ Params\end{tabular}} & \textbf{\begin{tabular}[c]{@{}c@{}}Generating \\ + \\ Training\\ Runtime \\ (Hours)\end{tabular}} & \textbf{\begin{tabular}[c]{@{}c@{}}Generating \\ + \\ Training\\ Energy \\ (kWh-PUE)\end{tabular}} & \textbf{\begin{tabular}[c]{@{}c@{}}Generating \\ + \\ Training\\ \ch{CO2} \\ Emission \\ (lbs)\end{tabular}} & \textbf{\begin{tabular}[c]{@{}c@{}}Inferencing \\ Frame \\ Per \\ Second\end{tabular}} & \textbf{\begin{tabular}[c]{@{}c@{}}Inferencing\\ Energy \\ Per Image\\ (kWh-PUE)\end{tabular}} & \textbf{\begin{tabular}[c]{@{}c@{}}Inferencing\\ \ch{CO2} \\ Emission \\ (lbs)\end{tabular}} \\ 

\midrule
& 1  &   27.43\%   &   1476536    &   6.26    &   0.6476    &  0.6178     &  13027.34     &  1.09e-05     &  1.04e-05                                                                                 \\ 
& 2  &  26.83\%    &   1476536    &   7.14    &   0.6950    &   0.6630    &   11856.32    &  7.75e-06     &  7.39e-06           
    \\ 
& 3  &   27.70\%   &   891448    &  7.21    &   0.7476    &  0.7132     &   13217.37    &    5.40e-06   &     5.16e-06        
    \\
& 4 &  17.83\%    &   116408    &   9.08    &  0.8835     &  0.8428     &  17062.41     &   2.71e-06    &   2.59e-06          
    \\
\multirow{-5}{*}{\textbf{\begin{tabular}[c]{@{}c@{}}GPT-4 \\ Turbo\end{tabular}}} & 5  &  25.23\% &  898104         &  9.47    &  1.01    &  0.96     &  15040.49     &   5.61e-06    &     5.35e-06              
    \\
\midrule
& 1 &  19.98\%    &  1113272     &  8.31     &   0.5679    &  0.5417     &   23322.39    &   8.85e-07   &  8.44e-07          
    \\
& 2 &  31.02\%    &   368440    &  6.53     &  0.4680     &   0.4468    &   186845.09    &   9.95e-07    & 9.49e-07            
    \\
& 3  &   24.75\%   &  35446424     &  11.30     &   1.707    &  1.638     &   8590.73   &   2.66e-06   &  2.54e-06            
    \\ 
& 4   &  23.52\%    &   21370    &   6.82    &  0.485     & 0.4630      &  22327.31     &   3.30e-06    &  3.15e-06           
    \\ 
\multirow{-5}{*}{\textbf{Gemini}}& 5& 22.49\%     &  2338776     &   4.99      &  0.32924     &  0.3141     &   21833.88    &  2.59e-06     & 2.47e-06            
    \\
\bottomrule
\end{tabular}
\end{sc}
\end{small}
\end{center}
\end{table*}
\begin{table*}[ht]
\begin{center}
\begin{small}
\begin{sc}
\caption{Comparison between LeMo-NADe and other NAS frameworks.}
\label{tab:comparison}
\renewcommand{\arraystretch}{1}
\small\addtolength{\tabcolsep}{2pt}
\begin{tabular}{lcccccccr}
\toprule
\multirow{2}{*}{\textbf{Method}} & \multicolumn{2}{c}{\textbf{CIFAR-10}} & \multicolumn{2}{c}{\textbf{CIFAR-100}} & \multicolumn{2}{c}{\textbf{ImageNet16-120}} \\
                                 & \textbf{Validation}  & \textbf{Test}  & \textbf{Validation}   & \textbf{Test}  & \textbf{Validation}     & \textbf{Test}     \\

\midrule
DARTS \cite{liu2018darts}            & 39.77 &  54.30 &  38.57            &  15.61              &  18.87           &  16.32           \\
DSNAS \cite{hu2020dsnas}           &  89.66 &  30.87  &  30.87 &  31.01 &  40.61  &  41.07   \\
PC-DARTS \cite{xu2019pc}        &  89.96  &  93.41   &  67.12  &  67.48  &  40.83 &  41.31  \\
iDARTS \cite{zhang2021idarts}          &  89.86  &  93.58  &  70.57  &  70.83   &  40.38    &  40.89  \\
GDAS \cite{dong2019searching}           &  89.89 &  93.61   &  71.34    &  70.70     &  41.59   &  41.71  \\
 $\beta$ -DARTS \cite{ye2022beta}         &  91.55   &  94.36   &  73.49   &  73.51        &  46.37           &  46.34           \\
 $\Lambda$ -DARTS \cite{movahedi2022lambda}         &  91.55           &  94.36           &  73.49            &  73.51              &  46.37           &  46.34           \\
GENIUS \cite{zheng2023can}          &  91.07  &  93.79  &  70.96   &  70.91   &  45.29   &  44.96   \\
\midrule
\textbf{LeMo-NADe (GPT4-Turbo)} &  90.90           &  89.41           &  68.38            &  67.90              &  27.05           &  27.70           \\
\textbf{LeMo-NADe (Gemini)}           &  82.94            &  81.76           &  52.12            &  52.96              &  30.34           &  31.02           \\

\bottomrule

\end{tabular}
\end{sc}
\end{small}
\end{center}
\end{table*}

\subsubsection{CIFAR-100 Dataset}

Table \ref{tab:cifar100} presents the outcomes of experiments performed with the CIFAR-100 dataset, following the same parameters outlined in Table \ref{tab:settings}. GPT-4 Turbo shows superior results regarding testing accuracy (67.90\%), lower model parameters (79530), lower generating + training energy (0.5021 kWh-PUE), and higher FPS (15079.86) for setting 4. 
The generated model has 64.55\% test accuracy when 0.3, 0.3, and 0.4 were provided as the priorities for FPS, training accuracy and validation accuracy respectively. For this model image inferencing took $7.44 \times 10^{-6}$ kWh-PUE energy.

Gemini performs better for setting 2 in terms of test accuracy, parameter, runtime and energy consumption. This generated model has an test accuracy of 52.96\%, took 2.48 hours for generating/training, and consumed 0.0889 kWh-PUE which is least among all other settings. Setting 3 leads to a test accuracy of 50.20\% where full priority is provided to the validation accuracy and it took 2.65 hours for generation/training.

\subsubsection{ImageNet16-120 Dataset}
Table \ref{tab:ImageNet} presents the outcomes of studies experiments carried out using the ImageNet16-120 dataset. LeMo-NADe with GPT-4 Turbo shows 27.70\% test accuracy for setting 3 which is the best accuracy among all settings. It takes 7.21 hours to generate and train the final model with an energy consumption of 0.7476 kWh-PUE. For setting 1 (0.4 priority on training accuracy and 0.6 priority on validation accuracy) we have a test accuracy of 27.43\% and a generation/training energy consumption of 0.6476 kWh-PUE.

Gemini also showed promising performance in terms of accuracy and model parameters, requiring low training/inferencing energy. As seen in Table \ref{tab:ImageNet}, LeMo-NADe using Gemini shows 31.02\% accuracy for setting 2 with comparatively lower model parameter (368440) having a training energy of 0.4680 kWh-PUE. For this dataset (ImageNet16-120) Gemini outperforms GPT-4 Turbo.

\subsubsection{Comparative Analysis}

Table \ref{tab:comparison} compares different leading-edge NAS techniques (where they used predefined search spaces to identify the optimal network) and LeMo-NADe (no search spaces) on CIFAR-10, CIFAR-100 and ImageNet16-120 datasets. We can see that for CIFAR-10 dataset, LeMo-NADe shows promising results and in some cases, it outperforms DARTS, DSNAS, PC-DARTS, iDARTS, and GDAS. For CIFAR-100 dataset, LeMo-NADe with GPT-4 Turbo outperforms DARTS, DSNAS, and PC-DARTS for both validation and test set results. For ImageNet16-120, LeMo-NADe with Gemini perform better than LeMo-NADe with GPT-4 Turbo and it shows 30.34\% and 31.02\% accuracy for validation and test set respectively which is approximately $2\times$ better than the DARTS method.

\section{In-depth Micro Analysis}
Fig. \ref{fig:gpt_gemini} shows the behavior of GPT-4 Turbo and Gemini based on the feedback from the Expert System. 

\subsection{GPT-4 Turbo}
\begin{figure*}[ht]
\begin{center}
\centerline{\includegraphics[width=0.95\textwidth]{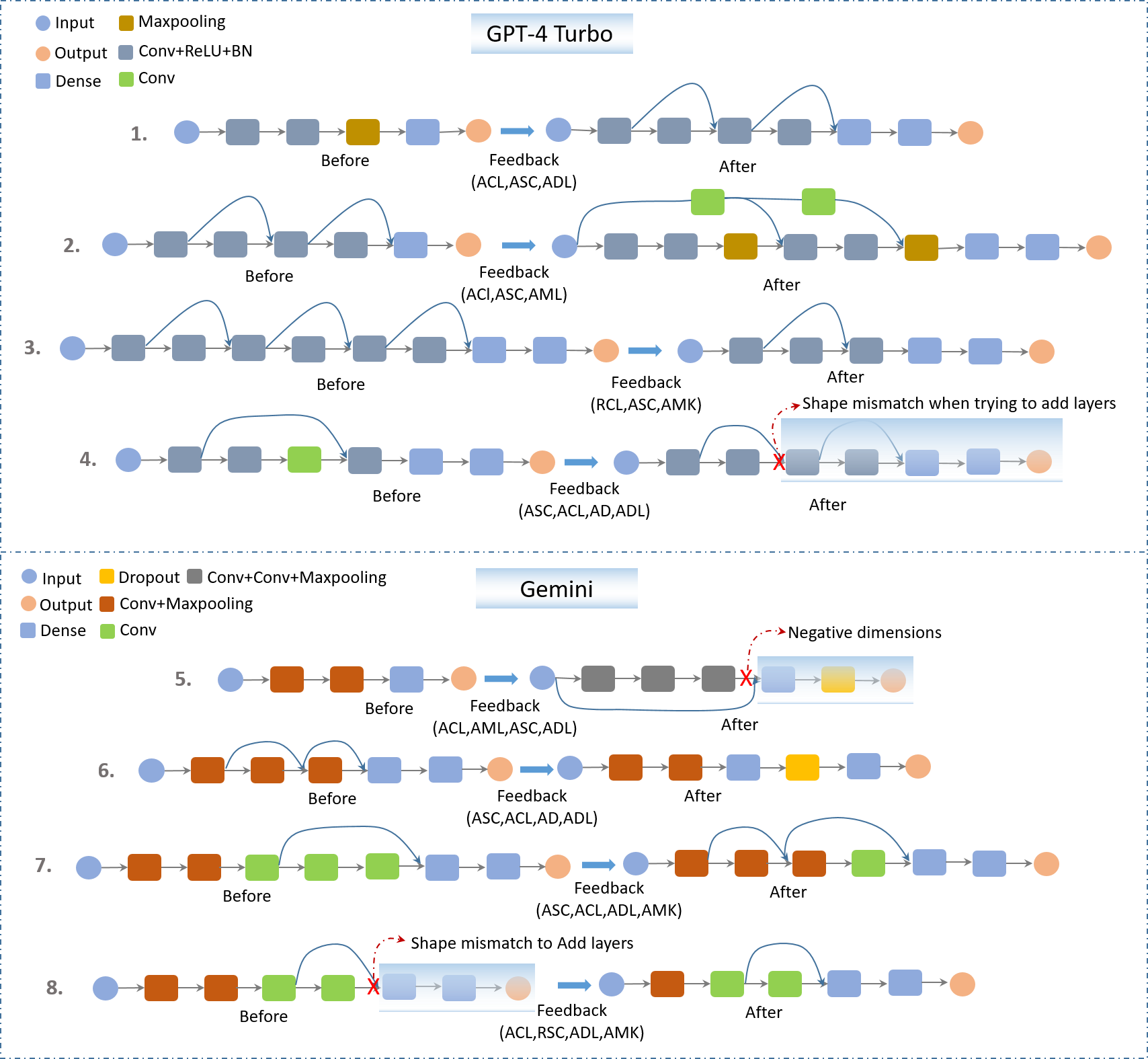}}
\caption{Qualitative analysis of the effectiveness of LeMo-NADe with GPT-4 Turbo and Gemini backends.}
\label{fig:gpt_gemini}
\end{center}
\end{figure*}

In Figure \ref{fig:gpt_gemini} we show how GPT-4 Turbo handles feedback from our expert system using the before and after neural network architecture representations. We describe individual cases below-

\begin{itemize}
    \item \textbf{Case 1}: We can see that the initially generated model was a simple CNN model with two convolutional layers having ReLU activations and Batch Normalization (BN), Maxpooling, and a single Dense layer. After getting the feedback (ACL,ASC,ADL) GPT-4 Turbo generated a model with four blocks of convolutional layers with ReLU activation and Batch Normalization (BN), and two Dense layers. It also adds two skip connections based on the feedback.

    \item  \textbf{Case 2}: Shows another behaviour of GPT-4 Turbo where it adds more skip connection by adding some convolutional layers. 
    
    \item \textbf{Case 3}: In some situations, skip connections make the network architecture more energy intensive to train/run. In this case, GPT-4 Turbo reduces the skip connections based on the feedback from ES.
    
    \item \textbf{Case 4}: GPT-4 Turbo not always show outstanding performance. We observe that approximately in 15\% of the cases it fails to follow the provided instructions. For example, in this case, it suggested a model with improper shape.

\end{itemize}

\subsection{Gemini}
Next, we analyze some interesting neural network generation scenarios when using LeMo-NADe powered by Gemini. 

\begin{itemize}
    \item \textbf{Case 5}: We can see that Gemini fails to build the network because of the negative dimensions. This happened because of using convolution layers with the maxpooling layers (having a larger filter size). 
    \item \textbf{Case 6}: We observe a behaviour where Gemini reduce the skip connection by taking the decision itself (it also describes the intuition behind this: \textit{reduce skip connection because accuracy doesn't improve by using this}) although LeMo-NADe suggested adding more skip connections, convolutional layers, and dense layers.
    \item \textbf{Case 7}: Gemini adds more convolutional layers with more skip connection as per the instructions given by LeMo-NADe.
    \item \textbf{Case 8}: Although Gemini shows promising results, in about 25\% of the cases it fails to generate a valid model because of negative dimensions and shape mismatches (when trying to concatenate with another layer). However, it tries to correct the error during the next iteration.

\end{itemize}

\section{Conclusion}
In this article, we formalized, implemented, and evaluated a multi-parameter neural discovery framework, LeMo-NADe that can generate novel neural networks for diverse requirements without having access to any pre-defined search space. The proposed framework operates with the help of a highly customizable metrics and rules driven expert system. This expert system is used to generate instructions for a backend Large Language Model (LLM) for iteratively generating different novel neural networks. LeMo-NADe was able to create highly effective (accuracy, FPS, power consumption) neural networks for different applications/requirements and for different datasets (CIFAR-10, CIFAR-100, and ImageNet16-120) in an efficient manner (generation time, energy, \ch{CO2} emissions). This work paves the way towards a new paradigm of AI-guided AI designing. Future works will explore the use of a customized LLM that is specifically trained to generate AI models for a wider range of user-defined applications.

\bibliography{example_paper}

\begin{thebibliography}{10}

\bibitem{chua2023tackling}
Michelle Chua, Doyun Kim, Jongmun Choi, Nahyoung~G Lee, Vikram Deshpande, Joseph Schwab, Michael~H Lev, Ramon~G Gonzalez, Michael~S Gee, and Synho Do.
\newblock Tackling prediction uncertainty in machine learning for healthcare.
\newblock {\em Nature Biomedical Engineering}, 7(6):711--718, 2023.

\bibitem{bhardwaj2023enhanced}
Rupali Bhardwaj and Ilina Tripathi.
\newblock An enhanced reversible data hiding algorithm using deep neural network for e-healthcare.
\newblock {\em Journal of Ambient Intelligence and Humanized Computing}, 14(8):10567--10585, 2023.

\bibitem{nandy2023intelligent}
Sudarshan Nandy, Mainak Adhikari, Venki Balasubramanian, Varun~G Menon, Xingwang Li, and Muhammad Zakarya.
\newblock An intelligent heart disease prediction system based on swarm-artificial neural network.
\newblock {\em Neural Computing and Applications}, 35(20):14723--14737, 2023.

\bibitem{jaafar2023multimodal}
Noussaiba Jaafar and Zied Lachiri.
\newblock Multimodal fusion methods with deep neural networks and meta-information for aggression detection in surveillance.
\newblock {\em Expert Systems with Applications}, 211:118523, 2023.

\bibitem{mahum2023robust}
Rabbia Mahum, Aun Irtaza, Marriam Nawaz, Tahira Nazir, Momina Masood, Sarang Shaikh, and Emad~Abouel Nasr.
\newblock A robust framework to generate surveillance video summaries using combination of zernike moments and r-transform and deep neural network.
\newblock {\em Multimedia Tools and Applications}, 82(9):13811--13835, 2023.

\bibitem{jan2023artificial}
Zohaib Jan, Farhad Ahamed, Wolfgang Mayer, Niki Patel, Georg Grossmann, Markus Stumptner, and Ana Kuusk.
\newblock Artificial intelligence for industry 4.0: Systematic review of applications, challenges, and opportunities.
\newblock {\em Expert Systems with Applications}, 216:119456, 2023.

\bibitem{raja2023industry}
Abirami Raja~Santhi and Padmakumar Muthuswamy.
\newblock Industry 5.0 or industry 4.0 s? introduction to industry 4.0 and a peek into the prospective industry 5.0 technologies.
\newblock {\em International Journal on Interactive Design and Manufacturing (IJIDeM)}, 17(2):947--979, 2023.

\bibitem{shafiq2023continuous}
Muhammad Shafiq, Kalpana Thakre, Kalluri~Rama Krishna, Noel~Jeygar Robert, Ashok Kuruppath, and Devendra Kumar.
\newblock Continuous quality control evaluation during manufacturing using supervised learning algorithm for industry 4.0.
\newblock {\em The International Journal of Advanced Manufacturing Technology}, pages 1--10, 2023.

\bibitem{rajput2023fault}
Dharmendra~Singh Rajput, Gaurav Meena, Malika Acharya, and Krishna~Kumar Mohbey.
\newblock Fault prediction using fuzzy convolution neural network on iot environment with heterogeneous sensing data fusion.
\newblock {\em Measurement: Sensors}, 26:100701, 2023.

\bibitem{liyakat2023machine}
Kazi Kutubuddin~Sayyad Liyakat.
\newblock Machine learning approach using artificial neural networks to detect malicious nodes in iot networks.
\newblock In {\em International Conference on Machine Learning, IoT and Big Data}, pages 123--134. Springer, 2023.

\bibitem{thakkar2023attack}
Ankit Thakkar and Ritika Lohiya.
\newblock Attack classification of imbalanced intrusion data for iot network using ensemble learning-based deep neural network.
\newblock {\em IEEE Internet of Things Journal}, 2023.

\bibitem{openai2023gpt}
R~OpenAI.
\newblock Gpt-4 technical report. arxiv 2303.08774.
\newblock {\em View in Article}, 2:13, 2023.

\bibitem{wang2023efficient}
Jianing Wang, Jinyu Hu, Yichen Liu, Zheng Hua, Shengjia Hao, and Yuqiong Yao.
\newblock El-nas: Efficient lightweight attention cross-domain architecture search for hyperspectral image classification.
\newblock {\em Remote Sensing}, 15(19):4688, 2023.

\bibitem{yang2023om}
Tiejun Yang, Qing He, and Lin Huang.
\newblock Om-nas: pigmented skin lesion image classification based on a neural architecture search.
\newblock {\em Biomedical Optics Express}, 14(5):2153--2165, 2023.

\bibitem{yang2023trustworthy}
Yi~Yang, Jiaxuan Wei, Zhixuan Yu, and Ruisheng Zhang.
\newblock A trustworthy neural architecture search framework for pneumonia image classification utilizing blockchain technology.
\newblock {\em The Journal of Supercomputing}, pages 1--34, 2023.

\bibitem{dong2023rd}
Peijie Dong, Xin Niu, Lujun Li, Zhiliang Tian, Xiaodong Wang, Zimian Wei, Hengyue Pan, and Dongsheng Li.
\newblock Rd-nas: Enhancing one-shot supernet ranking ability via ranking distillation from zero-cost proxies.
\newblock In {\em ICASSP 2023-2023 IEEE International Conference on Acoustics, Speech and Signal Processing (ICASSP)}, pages 1--5. IEEE, 2023.

\bibitem{wang2023dymc}
Jun Wang, Peng Yao, Feng Deng, Jianchao Tan, Chengru Song, and Xiaorui Wang.
\newblock Nas-dymc: Nas-based dynamic multi-scale convolutional neural network for sound event detection.
\newblock In {\em ICASSP 2023-2023 IEEE International Conference on Acoustics, Speech and Signal Processing (ICASSP)}, pages 1--5. IEEE, 2023.

\bibitem{li2023graph}
Jialin Li, Xuan Cao, Renxiang Chen, Xia Zhang, Xianzhen Huang, and Yongzhi Qu.
\newblock Graph neural network architecture search for rotating machinery fault diagnosis based on reinforcement learning.
\newblock {\em Mechanical Systems and Signal Processing}, 202:110701, 2023.

\bibitem{yuan2023ssob}
Wanqi Yuan, Chenping Fu, Risheng Liu, and Xin Fan.
\newblock Ssob: searching a scene-oriented architecture for underwater object detection.
\newblock {\em The Visual Computer}, 39(11):5199--5208, 2023.

\bibitem{jia2023fast}
Xiang Jia, Ying Tong, Hongming Qiao, Man Li, Jiangang Tong, and Baoling Liang.
\newblock Fast and accurate object detector for autonomous driving based on improved yolov5.
\newblock {\em Scientific reports}, 13(1):1--13, 2023.

\bibitem{mehta2023natural}
Ritik Mehta, Olha Jure{\v{c}}kov{\'a}, and Mark Stamp.
\newblock A natural language processing approach to malware classification.
\newblock {\em Journal of Computer Virology and Hacking Techniques}, pages 1--12, 2023.

\bibitem{girdhar2023benchmarking}
Nancy Girdhar, Micka{\"e}l Coustaty, and Antoine Doucet.
\newblock Benchmarking nas for article separation in historical newspapers.
\newblock In {\em International Conference on Asian Digital Libraries}, pages 76--88. Springer, 2023.

\bibitem{real2019regularized}
Esteban Real, Alok Aggarwal, Yanping Huang, and Quoc~V Le.
\newblock Regularized evolution for image classifier architecture search.
\newblock In {\em Proceedings of the aaai conference on artificial intelligence}, volume~33, pages 4780--4789, 2019.

\bibitem{liu2018progressive}
Chenxi Liu, Barret Zoph, Maxim Neumann, Jonathon Shlens, Wei Hua, Li-Jia Li, Li~Fei-Fei, Alan Yuille, Jonathan Huang, and Kevin Murphy.
\newblock Progressive neural architecture search.
\newblock In {\em Proceedings of the European conference on computer vision (ECCV)}, pages 19--34, 2018.

\bibitem{cai2018efficient}
Han Cai, Tianyao Chen, Weinan Zhang, Yong Yu, and Jun Wang.
\newblock Efficient architecture search by network transformation.
\newblock In {\em Proceedings of the AAAI Conference on Artificial Intelligence}, volume~32, 2018.

\bibitem{pham2018efficient}
Hieu Pham, Melody Guan, Barret Zoph, Quoc Le, and Jeff Dean.
\newblock Efficient neural architecture search via parameters sharing.
\newblock In {\em International conference on machine learning}, pages 4095--4104. PMLR, 2018.

\bibitem{liu2018darts}
Hanxiao Liu, Karen Simonyan, and Yiming Yang.
\newblock Darts: Differentiable architecture search.
\newblock {\em arXiv preprint arXiv:1806.09055}, 2018.

\bibitem{ying2019bench}
Chris Ying, Aaron Klein, Eric Christiansen, Esteban Real, Kevin Murphy, and Frank Hutter.
\newblock Nas-bench-101: Towards reproducible neural architecture search.
\newblock In {\em International conference on machine learning}, pages 7105--7114. PMLR, 2019.

\bibitem{dong2020bench}
Xuanyi Dong and Yi~Yang.
\newblock Nas-bench-201: Extending the scope of reproducible neural architecture search.
\newblock {\em arXiv preprint arXiv:2001.00326}, 2020.

\bibitem{krizhevsky2009learning}
Alex Krizhevsky, Geoffrey Hinton, et~al.
\newblock Learning multiple layers of features from tiny images.
\newblock 2009.

\bibitem{chrabaszcz2017downsampled}
Patryk Chrabaszcz, Ilya Loshchilov, and Frank Hutter.
\newblock A downsampled variant of imagenet as an alternative to the cifar datasets.
\newblock {\em arXiv preprint arXiv:1707.08819}, 2017.

\bibitem{ye2022beta}
Peng Ye, Baopu Li, Yikang Li, Tao Chen, Jiayuan Fan, and Wanli Ouyang.
\newblock $\beta$-darts: Beta-decay regularization for differentiable architecture search.
\newblock In {\em 2022 IEEE/CVF Conference on Computer Vision and Pattern Recognition (CVPR)}, pages 10864--10873. IEEE, 2022.

\bibitem{movahedi2022lambda}
Sajad Movahedi, Melika Adabinejad, Ayyoob Imani, Arezou Keshavarz, Mostafa Dehghani, Azadeh Shakery, and Babak~N Araabi.
\newblock $\lambda$ -darts: Mitigating performance collapse by harmonizing operation selection among cells.
\newblock {\em arXiv preprint arXiv:2210.07998}, 2022.

\bibitem{zheng2023can}
Mingkai Zheng, Xiu Su, Shan You, Fei Wang, Chen Qian, Chang Xu, and Samuel Albanie.
\newblock Can gpt-4 perform neural architecture search?
\newblock {\em arXiv preprint arXiv:2304.10970}, 2023.

\bibitem{achiam2023gpt}
Josh Achiam, Steven Adler, Sandhini Agarwal, Lama Ahmad, Ilge Akkaya, Florencia~Leoni Aleman, Diogo Almeida, Janko Altenschmidt, Sam Altman, Shyamal Anadkat, et~al.
\newblock Gpt-4 technical report.
\newblock {\em arXiv preprint arXiv:2303.08774}, 2023.

\bibitem{wang2023graph}
Haishuai Wang, Yang Gao, Xin Zheng, Peng Zhang, Hongyang Chen, and Jiajun Bu.
\newblock Graph neural architecture search with gpt-4.
\newblock {\em arXiv preprint arXiv:2310.01436}, 2023.

\bibitem{team2023gemini}
Gemini Team, Rohan Anil, Sebastian Borgeaud, Yonghui Wu, Jean-Baptiste Alayrac, Jiahui Yu, Radu Soricut, Johan Schalkwyk, Andrew~M Dai, Anja Hauth, et~al.
\newblock Gemini: a family of highly capable multimodal models.
\newblock {\em arXiv preprint arXiv:2312.11805}, 2023.

\bibitem{strubell2019energy}
Emma Strubell, Ananya Ganesh, and Andrew McCallum.
\newblock Energy and policy considerations for deep learning in nlp.
\newblock {\em arXiv preprint arXiv:1906.02243}, 2019.

\bibitem{hu2020dsnas}
Shoukang Hu, Sirui Xie, Hehui Zheng, Chunxiao Liu, Jianping Shi, Xunying Liu, and Dahua Lin.
\newblock Dsnas: Direct neural architecture search without parameter retraining.
\newblock In {\em Proceedings of the IEEE/CVF Conference on Computer Vision and Pattern Recognition}, pages 12084--12092, 2020.

\bibitem{xu2019pc}
Yuhui Xu, Lingxi Xie, Xiaopeng Zhang, Xin Chen, Guo-Jun Qi, Qi~Tian, and Hongkai Xiong.
\newblock Pc-darts: Partial channel connections for memory-efficient architecture search.
\newblock {\em arXiv preprint arXiv:1907.05737}, 2019.

\bibitem{zhang2021idarts}
Miao Zhang, Steven~W Su, Shirui Pan, Xiaojun Chang, Ehsan~M Abbasnejad, and Reza Haffari.
\newblock idarts: Differentiable architecture search with stochastic implicit gradients.
\newblock In {\em International Conference on Machine Learning}, pages 12557--12566. PMLR, 2021.

\bibitem{dong2019searching}
Xuanyi Dong and Yi~Yang.
\newblock Searching for a robust neural architecture in four gpu hours.
\newblock In {\em Proceedings of the IEEE/CVF Conference on Computer Vision and Pattern Recognition}, pages 1761--1770, 2019.

\end{thebibliography}
\bibliographystyle{unsrt}

\newpage
\appendix
\section{Appendix}

\begin{table*}[ht]
\centering

\begin{center}
\begin{small}
\begin{sc}
\caption{Effectiveness of LeMo-NADe for generating optimal neural network models for CIFAR-10 dataset using GPT-4 Turbo with different Temperature values.}
\label{tab:cifar10_appendix}
\renewcommand{\arraystretch}{1.2}
\scriptsize\addtolength{\tabcolsep}{-5pt}
\begin{tabular}{lccccccccr}
\toprule
\textbf{Temperature} & \textbf{Setting} & \textbf{\begin{tabular}[c]{@{}c@{}}Test \\ Accuracy\end{tabular}} & \textbf{\begin{tabular}[c]{@{}c@{}}Number\\ of \\ Params\end{tabular}} & \textbf{\begin{tabular}[c]{@{}c@{}}Generating \\ + \\ Training\\ Runtime \\ (Hours)\end{tabular}} & \textbf{\begin{tabular}[c]{@{}c@{}}Generating \\ + \\ Training\\ Energy \\ (kWh-PUE)\end{tabular}} & \textbf{\begin{tabular}[c]{@{}c@{}}Generating \\ + \\ Training\\ \ch{CO2} \\ Emission \\ (lbs)\end{tabular}} & \textbf{\begin{tabular}[c]{@{}c@{}}Inferencing \\ Frame \\ Per \\ Second\end{tabular}} & \textbf{\begin{tabular}[c]{@{}c@{}}Inferencing\\ Energy \\ Per Image\\ (kWh-PUE)\end{tabular}} & \textbf{\begin{tabular}[c]{@{}c@{}}Inferencing\\ \ch{CO2} \\ Emission \\ (lbs)\end{tabular}} \\ 

\midrule
 & 1  &  84.42\%   &   3491146    &   4.10    &   0.5545    &   0.5290    &    12009.58   &  7.50e-07     &  7.15e-06     \\ 
 & 2  &  85.06\%   &    621258   &   4.59    &    0.5794   &   0.5528    &   7461.53    &   5.08e-06    &  4.85e-06     \\
 & 3  &  85.65\%   &   3494730    &   4.41    &   0.8478    &   0.8088    &    10904.14   &   7.15e-06    &   6.82e-06    \\
 & 4  &  66.25\%   &   795330    &   4.75    &   0.4531    &   0.4323    &   15335.26    &  5.28e-06     &  5.04e-06     \\
 
\multirow{-5}{*}{T = 0.2} & 5  &  87.32\%   &   4605642    &    4.05   &    0.5374   &  0.5127     &   13167.46    &  6.09e-06     &   5.81e-06    \\
\midrule
& 1  &  83.37\%   &   3491146    &   3.99    &   0.6099    &  0.5819     &   11487.42    &  1.09e-06     &  1.04e-06     \\ 
 & 2  &  84.03\%   &   621258    &   4.08    &   0.5011    &   0.4781    &   13521.36    &   5.66e-06    &  5.40e-06     \\
 & 3  &  87.59\%   &   3597002    &   3.89    &   0.6717    &  0.6408     &  108132.38     &   7.24e-06    &   6.91e-06    \\
 & 4  &  65.65\%   & 79530      &  4.70    &   0.5086    &   0.4852    &  15032.04    &   5.93e-06    &  5.66e-06     \\
\multirow{-5}{*}{T = 0.4} & 5  &  89.02\%   &   3418570    &   4.27    &  0.5725     &   0.5461    & 10377.16   &  1.09e-05    &   1.04e-05     \\

\midrule
& 1  &  89.70\%   &   3428938    &  3.92     &   0.5068    &  0.4834     &  11013.01     &   1.03e-05    &   9.81e-06    \\ 
 & 2  &  84.11\%   &   620746    &   4.15    &   0.4889    &   0.4664    &   12560.41    &   5.02e-06    &  4.79e-06     \\
 & 3  &  90.90\%   &    5884042   &   4.05    &   0.7806    &   0.7447    &   9218.27    &   7.41e-06    &   7.07e-06    \\
 & 4  &  65.14\%   &   79530    &   4.65    &   0.4436    &   0.4232    &   15426.29    &  2.63e-06     &  2.51e-06     \\
\multirow{-5}{*}{T = 0.6} & 5  &  88.67\%   &    349530   &   4.20    &  0.5266     &   0.5024    &   11622.50    &  9.99e-06     &  9.54e-06     \\

\midrule
& 1  &  89.13\%   &   4997898    &   4.84    &   2.09    &  2.0     &  4743.40     &   1.31e-05    &  1.25e-05     \\ 
 & 2  &  83.28\%   &   361930    &   3.57    &  0.4518     &  0.4301     &   12494.94    &   5.54e-06    &  5.28e-06     \\
 & 3  &  89.53\%   &   4778762    &    3.99   &   0.6022    &  0.5745     &  10870.27     &   6.61e-06    &   6.30e-06    \\
 & 4  &  63.48\%   &   231562    &   4.66    &  0.4759     &  0.4541     &  15866.69     &  2.78e-06     &  2.66e-06     \\
\multirow{-5}{*}{T = 0.8} & 5  &  90.46\%   &  9675082     &   3.61    &  0.4646     &   0.4432    &   11438.98    &    1.01e-05   &  9.60e-06     \\

\midrule
& 1  &  82.22\%   &   2401866    &   3.67    &   0.4656    &  0.4442     &  11855.67     &   9.06e-06    &  8.63e-06     \\ 
 & 2  &  86.34\%   &  9188298     &  4.24     &   0.5778    &   0.5512    &  12245.07     &   6.56e-06    &  6.26e-06     \\
 & 3  &  84.55\%   &  3510858     &   3.78    &  0.5030     &  0.4798     &   12511.31    &  6.82e-06     &  6.50e-06     \\
 & 4  & 67.17\%    &   79530    &   4.69    &   0.4811    &  0.4589     &  14869.92     &  5.89e-06     &   5.62e-06    \\
\multirow{-5}{*}{T = 1} & 5  &  83.15\%   &   161098    &   3.77    &  0.4118     &  0.3929     &  14786.20     &  5.85e-06     &   5.58e-06

\\ 
\bottomrule

\end{tabular}
\end{sc}
\end{small}
\end{center}
\end{table*}
\begin{table*}[ht]
\centering
\begin{center}
\begin{small}
\begin{sc}
\caption{Effectiveness of LeMo-NADe for generating optimal neural network model for CIFAR-100 dataset using GPT-4 Turbo with different Temperature values.}
\label{tab:cifar100_appendix}
\renewcommand{\arraystretch}{1}
\scriptsize\addtolength{\tabcolsep}{-5pt}
\begin{tabular}{lccccccccr}
\toprule
\textbf{Temperature} & \textbf{Setting} & \textbf{\begin{tabular}[c]{@{}c@{}}Test \\ Accuracy\end{tabular}} & \textbf{\begin{tabular}[c]{@{}c@{}}Number\\ of \\ Params\end{tabular}} & \textbf{\begin{tabular}[c]{@{}c@{}}Generating \\ + \\ Training\\ Runtime \\ (Hours)\end{tabular}} & \textbf{\begin{tabular}[c]{@{}c@{}}Generating \\ + \\ Training\\ Energy \\ (kWh-PUE)\end{tabular}} & \textbf{\begin{tabular}[c]{@{}c@{}}Generating \\ + \\ Training\\ \ch{CO2} \\ Emission \\ (lbs)\end{tabular}} & \textbf{\begin{tabular}[c]{@{}c@{}}Inferencing \\ Frame \\ Per \\ Second\end{tabular}} & \textbf{\begin{tabular}[c]{@{}c@{}}Inferencing\\ Energy \\ Per Image\\ (kWh-PUE)\end{tabular}} & \textbf{\begin{tabular}[c]{@{}c@{}}Inferencing\\ \ch{CO2} \\ Emission \\ (lbs)\end{tabular}} \\ 

\midrule
 & 1  &  56.43\%   &  3489316     &    4.50   &  0.8881     &  0.8472     &    9389.76   &  1.07e-05     &  1.02e-05     \\ 
 & 2  &  58.03\%   &  948708     &   4.18    &  0.5707    &  0.5444     &   12694.01    &   5.11e-06    &   4.88e-06    \\
 & 3  &   58.85\%  &  948708     &  4.19     &  0.8080     &   0.7709    &   12374.44    &   5.72e-06    &  5.46e-06     \\
 & 4  &   45.19\%  &  635172     &   4.77    &  0.4730     &   0.4512    &   15020.50    &   5.35e-06    &   5.11e-06    \\
\multirow{-5}{*}{T = 0.2} & 5  &  57.67\%   &    948708   &  4.57     &   0.7865    &   0.7503    &  13518.15     &   5.32e-06    &  5.07e-06     \\
\midrule
& 1  &  57.74\%   &   948708    &    4.21   &   0.7334    &  0.6997     &  11285.20     &  5.38e-06     &  5.13e-06     \\ 
 & 2  &  62.42\%   &   463380    &    4.25   &   0.6635    & 0.6330      &   11060.66    &   6.19e-06    &   5.90e-06    \\
 & 3  &  53.45\%   &  4630308     &  4.26     &   0.7017    &  0.6694     &  10563.94     &  5.64e-06     &  5.38e-06     \\
 & 4  &  45.28\%   &  635172     &  4.71     &   0.4597    &  0.4385     &   14895.69    &   7.96e-06    &  7.59e-06     \\
\multirow{-5}{*}{T = 0.4} & 5  &  61.90\%   &   34166180    &   4.62    &  0.8843     &   0.8436    &  11405.19     &   1.08e-05    &  1.03e-05     \\

\midrule
& 1  &  56.22\%   &   1543844    &  3.95     &  0.6204     &  0.5919     &  9703.77     &  7.62e-06     &   7.27e-06    \\ 
 & 2  &  59.34\%   &   948708    &   4.05    &   0.6515    &  0.6215     &   13147.95    &   5.96e-06    &   5.69e-06    \\
 & 3  &  44.92\%   &   6108708    &   4.38    &   0.7732    &   0.7377    &   11135.41    &  6.73e-06     &  6.42e-06     \\
 & 4  &  46.52\%   &    635172   &   4.74    &   0.4518    &   0.4310    &   14760.20    &   5.12e-06    &  4.89e-06     \\
\multirow{-5}{*}{T = 0.6} & 5  &   60.44\%  &  3441700     &   3.99    &   0.6337    &  0.6046     &   12014.46    &   7.04e-06    &  6.71e-06     \\

\midrule
& 1  &  54.60\%   &   4624932    &  3.99     &  0.5776     &  0.5510     &   11605.52    &  5.69e-06     & 5.43e-06      \\ 
 & 2  &  63.69\%   &  2674852     &  4.42     &   0.6246    &   0.5958    &   12606.06    &  5.73e-06     &  5.46e-06     \\
 & 3  &  50.70\%   &  3557284     &  3.88     &  0.6374     &  0.6080     &  12400.02     &   1.04e-05    &  9.92e-06     \\
 & 4  &  40.80\%   &   325956    &  4.72     &  0.4644     &   0.4431    &   14583.39    &  2.61e-06     &  2.49e-06     \\
\multirow{-5}{*}{T = 0.8} & 5  &  67.52\%   &  6009124     &  4.36     &   0.6583    &  0.6280     &  10632.62     &   1.10e-05    &  1.05e-05     \\

\midrule
& 1  &  51.60\%   &  6166756     &   3.89    &  0.5683     &   0.5421    &   11330.60    &  6.53e-06     &  6.23e-06     \\ 
 & 2  &  56.28\%   &  948708     &   3.76    &   0.4831    &    0.4609   &  13364.18     &  5.26e-06     &  5.02e-06     \\
 & 3  &  56.27\%   &  3489316     &  4.17     &   0.6477    &  0.6179     &   9965.78    &  1.05e-05     & 1.00e-06      \\
 & 4  &  40.05\%   &   325956    &  4.72     &  0.4527     &  0.4318     &   14878.67    &  5.14e-06     &  4.91e-06     \\
\multirow{-5}{*}{T = 1} & 5  &  60.75\%   &  92465646     &   4.18    &   0.5933    &  0.5660     &  12972.91     &  6.11e-06     &   5.82e-06

\\ 
\bottomrule

\end{tabular}
\end{sc}
\end{small}
\end{center}
\vskip -0.1in
\end{table*}
\begin{table*}[ht]
\centering
\begin{center}
\begin{small}
\begin{sc}
\caption{Effectiveness of LeMo-NADe for generating optimal neural network models for ImageNet16-120 dataset using GPT-4 Turbo with different Temperature values.}
\label{tab:imagenet_appendix}
\renewcommand{\arraystretch}{1.2}
\scriptsize\addtolength{\tabcolsep}{-5pt}
\begin{tabular}{lccccccccr}
\toprule
\textbf{Temperature} & \textbf{Setting} & \textbf{\begin{tabular}[c]{@{}c@{}}Test \\ Accuracy\end{tabular}} & \textbf{\begin{tabular}[c]{@{}c@{}}Number\\ of \\ Params\end{tabular}} & \textbf{\begin{tabular}[c]{@{}c@{}}Generating \\ + \\ Training\\ Runtime \\ (Hours)\end{tabular}} & \textbf{\begin{tabular}[c]{@{}c@{}}Generating \\ + \\ Training\\ Energy \\ (kWh-PUE)\end{tabular}} & \textbf{\begin{tabular}[c]{@{}c@{}}Generating \\ + \\ Training\\ \ch{CO2} \\ Emission \\ (lbs)\end{tabular}} & \textbf{\begin{tabular}[c]{@{}c@{}}Inferencing \\ Frame \\ Per \\ Second\end{tabular}} & \textbf{\begin{tabular}[c]{@{}c@{}}Inferencing\\ Energy \\ Per Image\\ (kWh-PUE)\end{tabular}} & \textbf{\begin{tabular}[c]{@{}c@{}}Inferencing\\ \ch{CO2} \\ Emission \\ (lbs)\end{tabular}} \\ 

\midrule
 & 1  &  30.25\%   &   2467128    &   6.40    &  0.708     & 0.6755   &   11737.36    &  8.0e-06    &  7.64e-06     \\ 
 & 2  &  24.22\%   &  429656     &  6.20     &  0.6340     & 0.6049   &  13649.14     &  5.51e-06    &  5.25e-06     \\ 
 & 3  &  26.50\%   &   3296760    &  7.44     & 0.8343      & 0.7960   &   15027.43    &  8.78e-06    &  8.37e-06     \\ 
 & 4 & 18.40\%    &  116408     &  9.35     &   0.8622    & 0.8225   &   15690.88    &  5.12e-06    &  4.88e-06     \\ 
\multirow{-5}{*}{T = 0.2} & 5  &  25.57\%  & 881592 &  8.65     & 0.8700      &  0.8300     &  15686.66  &  5.35e-06     &  5.10e-06     \\ 
\midrule
& 1  &  25.97\%    &  2686968   &  7.03     &  0.7098     & 0.6772      &  15062.65  & 8.02e-06      &   7.65e-06       \\ 
 & 2    &  26.07\%   &  1473208     &   7.74    &  0.7508     & 0.7175   &   14646.19    &  5.14e-06 & 4.90e-06       \\ 
 
 & 3  &  29.08\%   &  901688     &  6.85     &   0.6782    &  0.6470  &   12419.29    &  5.13e-06 & 4.89e-06       \\ 
 & 4  &  14.73\%   &   68952    &   8.99    &   0.8518    &  0.8126  &   18408.92    &  2.64e-06 & 2.52e-06        \\ 
\multirow{-5}{*}{T = 0.4} & 5  &  26.50\%   &   1488184    &  8.49     &  0.9066     &  0.8649     &   14915   &   5.70e-06   &   5.44e-06    \\ 

\midrule
& 1  &  2.635\%    &  1492792   &  6.98     & 0.7321      & 0.6984      & 12709.54   &  5.37e-06     &  5.12e-06        \\ 
 & 2  & 27.73\%       &  899640     &   7.34    &  0.7384     & 0.7044   &  13815.50     &  5.26e-06  & 5.02e-06       \\
 & 3  &  25.37\%   &  883128     &  5.94     &  0.5869     & 0.5599   &  15487.06     &  5.18e-06   &   4.94e-06       \\ 
 & 4  &   19.40\%  &   212536    &   8.49    &   0.8620    &  0.8223  &  18041.92     &  5.65e-06   &   5.39e-06        \\ 
\multirow{-5}{*}{T = 0.6} & 5  &  23.72\%   &   898104    &  8.05     &  0.8143     &  0.7768     &  14988.14    &  5.41e-06    &   5.17e-06     \\ 

\midrule
& 1  &  35.03\%    &  1442744   &  8.05     &   0.9968    &  0.9510     &  12202.11  &   6.55e-06    &   6.25e-06       \\ 
 & 2    &  26.02\%   &  2064056     &  7.85     &  0.8090     & 0.7718   &   12326.82    &   5.42e-06   &   5.17e-06      \\ 
 & 3  &  27.42\%   &  1843192     & 6.75      &  0.7034     & 0.6710   &  11216.21     &  5.38e-06  &   5.13e-06       \\ 
 & 4  &  17.73\%   &  116024     &  8.77     &  0.8727     &  0.8325  &  19443.04     & 5.51e-06    &   5.26e-06         \\ 
\multirow{-5}{*}{T = 0.8} & 5  & 26.73\%    &  832312     &   6.16    &  0.5988     &   0.5712    &  15291.25    &   5.07e-06   &   4.84e-06    \\ 

\midrule
& 1  &  26.17\%    &  4625464   &  6.74     &  0.6879     &  0.6562     & 16176.14   &  5.31e-06     &   5.06e-06       \\ 
 & 2    &  28.12\%   &  418456     &   6.25    &   0.6454    &  0.6157  &  12878.29     &  5.47e-06 &   5.23e-06       \\ 
 & 3  &  27.93\%   &  111160     &  7.17     &  0.7335     & 0.6997   &   14035.10    &  5.23e-06   &   4.99e-06       \\ 
 & 4  &  10.13\%   &   38024    &   8.92    &   0.8566    & 0.8173   & 17140.02      &   5.38e-06   &   5.13e-06       \\ 
\multirow{-5}{*}{T = 1} & 5  &  31.65\%   &  3294584     &   7.16    &    0.7576   &   0.7227    &  15842.34    &  5.38e-06 &   5.14e-06     \\

\bottomrule

\end{tabular}
\end{sc}
\end{small}
\end{center}
\end{table*}

Table \ref{tab:cifar10_appendix}, \ref{tab:cifar100_appendix}, \ref{tab:imagenet_appendix} shows the experimental results of LeMo-NADe when using GPT-4 Turbo on CIFAR-10, CIFAR-100 and ImageNet16-120 dataset respectively. We varied the temperature within the range of 0.2 to 1 and evaluated the generated neural network architecture. After carefully investigating the results we can see in Table \ref{tab:cifar10_appendix} that LeMo-NADe perform better for Temperature 0.6 in terms of accuracy, model parameter, computing energy and inference FPS. On the other hand, in Table \ref{tab:cifar100_appendix} we can see that LeMo-NADe with GPT-4 Turbo shows better performance for Temperature value 0.8. It also shows better results for Temperature 0.8 for ImageNet16-120 dataset as reported in Table \ref{tab:imagenet_appendix}.

\begin{figure}[ht]
\begin{center}
\includegraphics[width=0.50\linewidth]{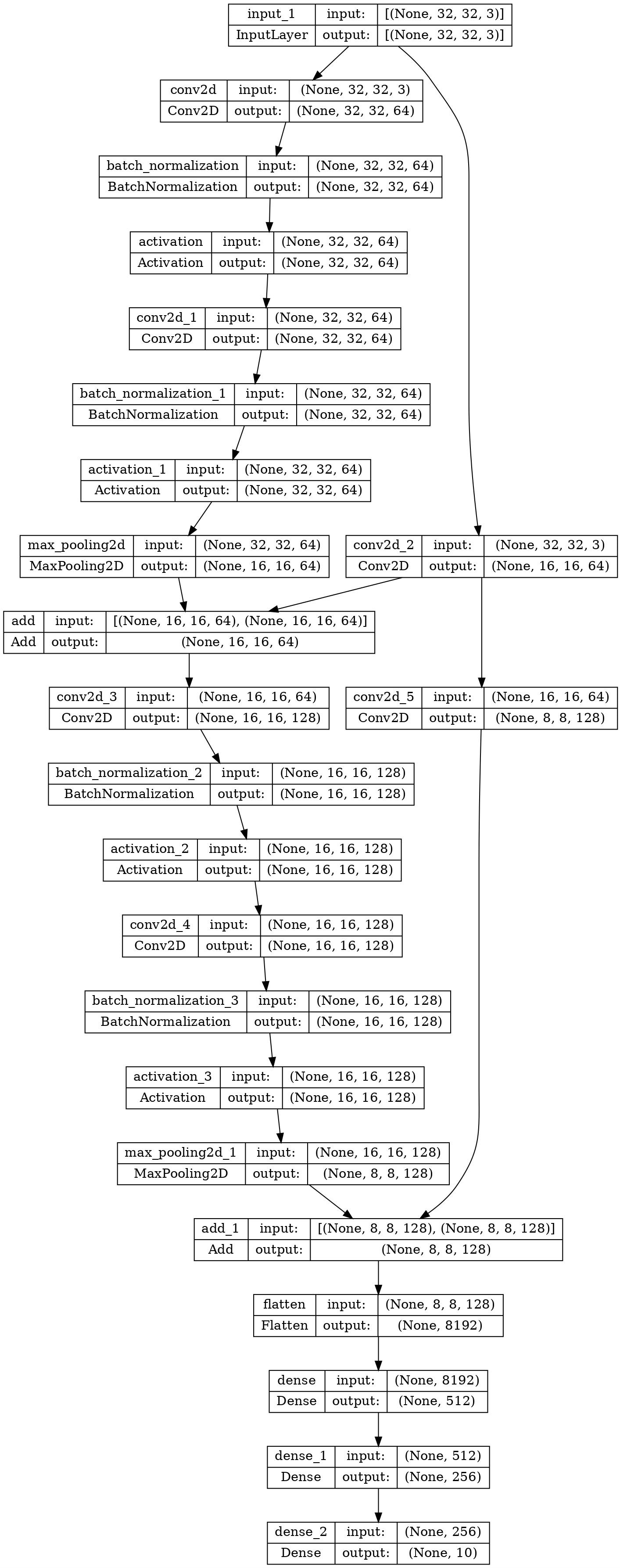}
\caption{Model generated by GPT-4 Turbo for CIFAR-10 dataset with temperature 0.6 and setting 3, and it gives 90.90\% test accuracy.}
\label{fig:cifar10_gpt}
\end{center}
\end{figure}

Figure \ref{fig:cifar10_gpt} illustrates a model generated by LeMo-NADe with GPT-4 Turbo for CIFAR-10 dataset. This model is obtained for the following configuration-

Temperature = 0.6; Priorities: Training accuracy: 0, Training energy: 0, Validation accuracy: 1, Validation energy: 0, FPS: 0; Threshold: Training accuracy: 0.95, Validation accuracy: 0.92, FPS: 20000, Training energy: $10^{-3}$ kWh, Validation energy: $10^{-5}$ kWh.

\begin{figure*}[ht]
\begin{center}
\includegraphics[width=0.50\linewidth]{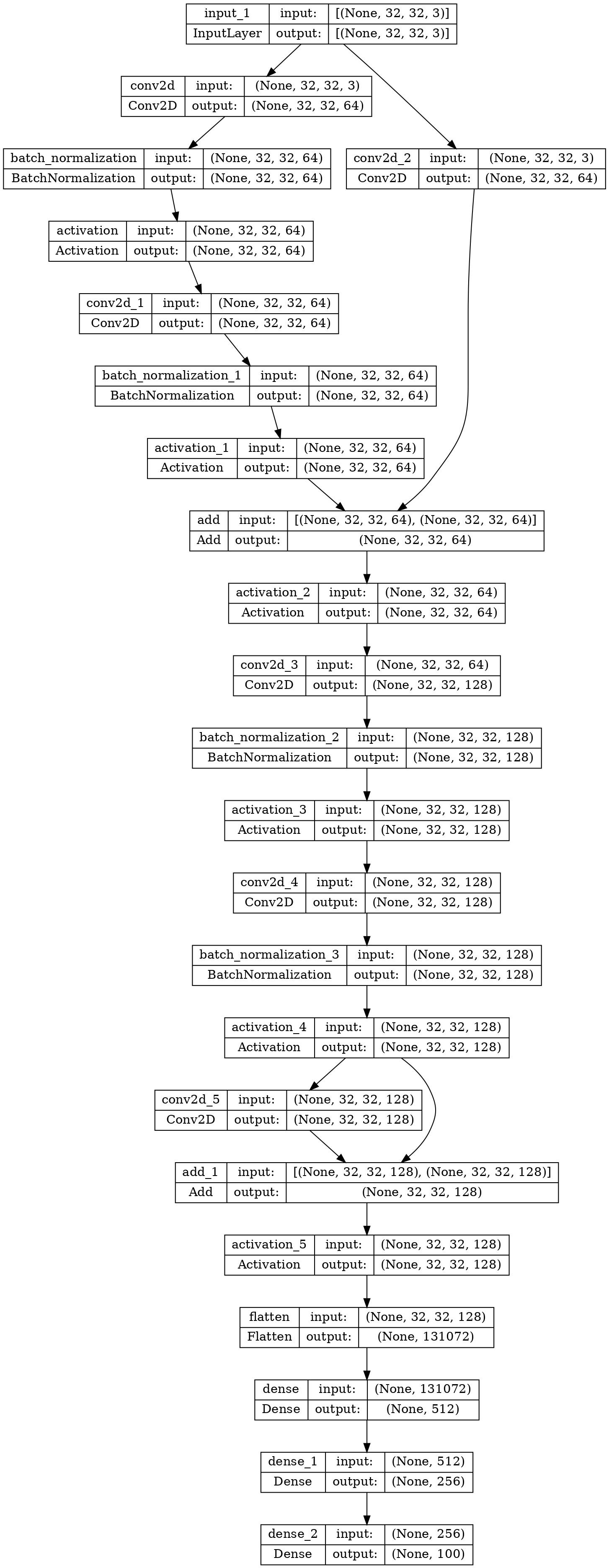}
\caption{Model generated by GPT-4 Turbo for CIFAR-100 dataset with temperature 0.4 and setting 1, and it gives 57.74\% test accuracy.}
\label{fig:cifar100_gpt}
\end{center}
\end{figure*}

Figure \ref{fig:cifar100_gpt} illustrates a model generated by LeMo-NADe with GPT-4 Turbo for CIFAR-100 dataset. This model is obtained for the following configuration-

Temperature = 0.4; Priorities: Training accuracy: 0.4, Training energy: 0, Validation accuracy: 0.6, Validation energy: 0, FPS: 0; Threshold: Training accuracy: 0.90, Validation accuracy: 0.85, FPS: 20000, Training energy: $10^{-2}$ kWh, Validation energy: $10^{-4}$ kWh.

\begin{figure*}[ht]
\begin{center}
\includegraphics[width=0.50\linewidth, height= 0.98\textheight]{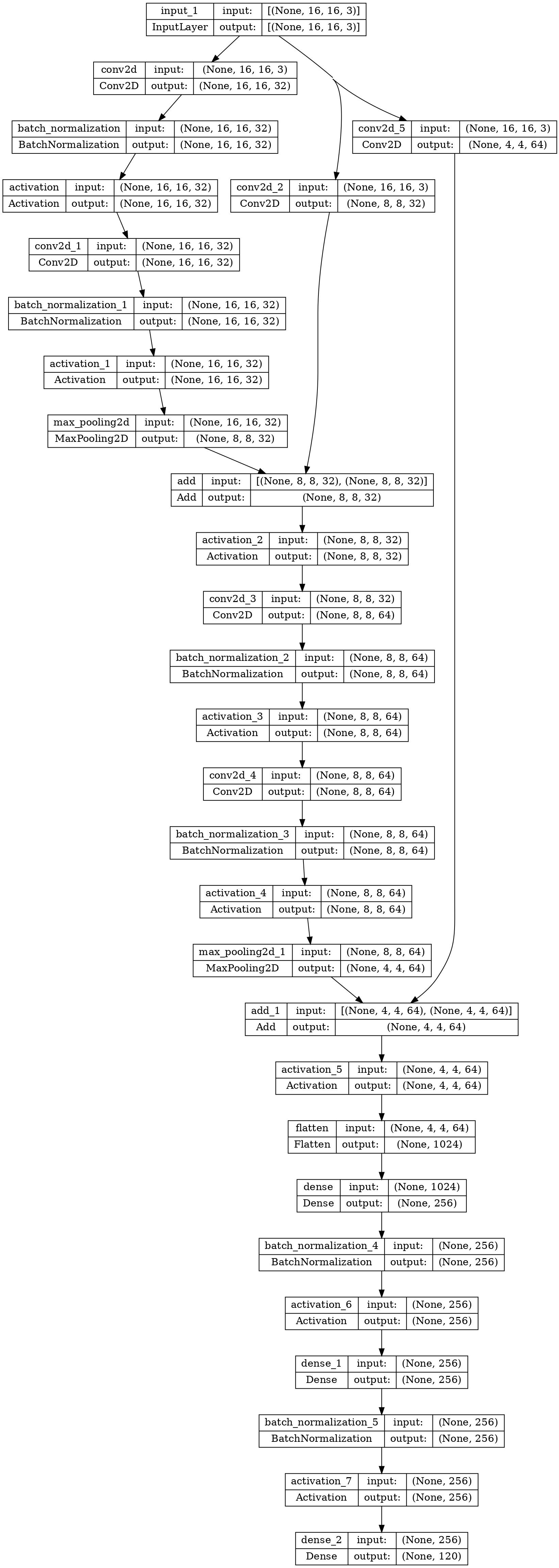}
\caption{Model generated by GPT-4 Turbo for ImageNet16-120 dataset with temperature 0.4 and setting 1, and it gives 25.97\% test accuracy.}
\label{fig:image16-120}
\end{center}
\end{figure*}

Figure \ref{fig:image16-120} illustrates a model generated by LeMo-NADe with GPT-4 Turbo for ImageNet16-120 dataset. This model is obtained for the following configuration-

Temperature = 0.4; Priorities: Training accuracy: 0.4, Training energy: 0, Validation accuracy: 0.6, Validation energy: 0, FPS: 0; Threshold: Training accuracy: 0.70, Validation accuracy: 0.65, FPS: 22000, Training energy: $10^{-3}$ kWh, Validation energy: $10^{-5}$ kWh.

\end{document}